# Active management of battery degradation in wireless sensor network using deep reinforcement learning for group battery replacement

Jong-Hyun Jeong[a], Hongki Jo[a*], Qiang Zhou[b], Tahsin Afroz Hoque Nishat[a], Lang Wu[c]

[a]*Department of Civil & Architectural Engineering & Mechanics, University of Arizona, Tucson, AZ, USA*

[b]*Department of Systems & Industrial Engineering, University of Arizona, Tucson, AZ, USA*

[c] *Shenzhen Institutes of Advanced Technology Chinese Academy of Sciences, Shenzhen, Guangdong, CN*

**Abstract**

Wireless sensor networks (WSNs) have become a promising solution for structural health monitoring (SHM), especially in hard-to-reach or remote locations. Battery-powered WSNs offer various advantages over wired systems, however limited battery life has always been one of the biggest obstacles in practical use of the WSNs, regardless of energy harvesting methods. While various methods have been studied for battery health management, existing methods exclusively aim to extend lifetime of individual batteries, lacking a system level view. A consequence of applying such methods is that batteries in a WSN tend to fail at different times, posing significant difficulty on planning and scheduling of battery replacement trip. This study investigate a deep reinforcement learning (DRL) method for active battery degradation management by optimizing duty cycle of WSNs at the system level. This active management strategy effectively reduces earlier failure of battery individuals which enable group replacement without sacrificing WSN performances. A simulated environment based on a real-world WSN setup was developed to train a DRL agent and learn optimal duty cycle strategies. The performance of the strategy was validated in a long-term setup with various network sizes, demonstrating its efficiency and scalability.

## 1. Introduction

The growth of the Internet of Things (IoT) has led to a significant number of battery-powered wireless devices to be used in the field. IoT devices can easily allow access for the automation and remote control of a wide range of tasks. So, it can revolutionize many industries by increasing efficiency and enabling the collection and analysis of large amounts of data. However, limited battery life has been one of the biggest obstacles for wireless devices to be widely used in many IoT applications in practice. In 2019, the Global System for Mobile Communications Association (GSMA) has predicted the number of IoT devices to reach 25 billion in the year 2025 [1]. If half of these devices are battery-powered and assume a 5-year battery life, then there will be around 5 billion batteries being replaced each year. This means that there is a desperate need for cost-effective battery management strategies for replacement decision-making. Wireless sensor networks (WSNs) have emerged as a popular trend in IoT applications, particularly in the field of structural health monitoring (SHM). These are especially seen in large-scale civil

———
[*] Corresponding author.; e-mail:hjo@arizona.edu.

infrastructures like bridges, buildings, offshore wind turbines, and power plant systems. These WSNs are considered a cost-effective solution for SHM, which makes monitoring and real-time data collection much easier to ensure the structural integrity and safety of these critical infrastructure systems [2].

In such WSNs for SHM, effective power and battery management is a key in the success of, particularly, long-term operation of WSNs. No matter what battery technologies are employed, service life of a battery is limited, requiring the replacement to continue the desired service of WSNs. Particularly, the WSNs for long-term SHM of infrastructure are often installed in remote locations, such as bridges in mountainous areas, oil pipelines in deserts, and offshore wind turbines. In these scenarios, the cost of planning and logistics of maintenance trips can be very weighty. Therefore, it is not economically feasible to conduct maintenance trips for replacing a single or a few batteries that reached their end-of-life. Instead, replacing the batteries as a group during a single maintenance trip is often required for these scenarios.

Numerous studies have investigated ways to extend the lifespan of solar-powered WSN nodes. One approach involves using hybrid energy storage systems, which typically combine batteries and supercapacitors [3–5]. Additionally, researchers have explored techniques like biased partial charging prioritization [6], advanced power consumption modeling in the IEEE 802.11 standard [7], and cutoff voltage control to prevent nonlinear degradation, all aimed at improving the battery lifespan of WSN [8].

However, existing methods for battery health management exclusively aim to extend the lifetime of individual batteries, lacking a system level view. A consequence of applying such conventional battery management strategies is that batteries in the WSN tend to fail at very different times, posing significant difficulty on planning and scheduling group battery replacement activities.

In addition to power management, ensuring the desired quality of service is also an essential factor in the success of WSNs for SHM. To ensure the service quality, maximizing the system utility (i.e., number of active nodes) is critical. This is especially important for SHM applications, because the more number of sensors are used in monitoring structural responses, the more comprehensive information can be obtained about the structure's health conditions. Considering the majority of WSN systems for SHM employ a power conserving scheme that utilizes sleep and idle modes, careful planning of active node assignments is needed to ensure that as many nodes as possible can be reached upon the detection of an event without sacrificing the power saving effort.

A notable amount of research has been conducted to manage the energy efficiency and service quality for WSN applications. For example, optimization-based approaches have been investigated to find the optimal number of sensor nodes and network layout [6,9]. One popular protocol that aims to extend the lifetime of WSN systems is the Low-Energy Adaptive Clustering Hierarchy (LEACH). This protocol can distribute the energy load evenly throughout the system [10]. Subsequent developments of LEACH have been proposed to consider energy harvesting [11] and duty cycle scheduling for energy neutral operation (ENO) that makes total amount of harvested energy match with the consumed energy of each node [12].

While a number of ENO related studies have been developed, reinforcement learning (RL) based approaches have particularly been studied as a flexible and powerful method. Among them, several studies have utilized tabular RL algorithms such as Q-learning and SARSA to achieve ENO for individual nodes [13,14]. However, these approaches have not considered the system-level service quality, but only focus on the node-level performance. In particular, SHM applications require system-level centralized control of the WSN. To ensure the proper number of nodes are active when needed to detect important events and to achieve desired quality of SHM performances, system-level control is needed. Furthermore, these tabular RL-based approaches are limited in handling large-scale networks and can only be used for small-scale WSNs.

A recent study by Long and Buyukozturk [15] has proposed a network-level WSN duty cycle control strategy using deep reinforcement learning (DRL). They proposed a centralized control approach for a WSN in order to maximize the system's overall utility. They took into account various practical uncertainties, including problems in solar harvesting from real-life irradiance and meteorological data, as well as in the geometric wireless connectivity between nodes. However, the performance has been validated with a small WSN comprised of only 16 nodes arranged in a 2D square lattice configuration and long-term battery degradation was not considered.

Up to now, most of the prior advancements have been primarily focused on enhancing the energy management of individual nodes or devising strategies for system utility planning. And they employed small WSNs for validation purposes, which could limit their scalability in real-world scenarios. No study has been done to comprehensively consider the long-term battery degradation at system level, system utility planning, and scalability at the same time. To bridge this gap, the present study introduces a DRL strategy for active management of battery degradation, which allows uniform battery degradation and enable group replacement, while maximizing the system utility. And the performance is validated with various-size real-world WSN configuration. This approach has the following objectives:

i) Maximize the system utility within practical WSN setups, considering real-world uncertainties by employing deep reinforcement learning (DRL).

ii) Address the long-term battery degradation issues at the system level, treating the problem as a sequential decision-making challenge within the realm of reinforcement learning (RL).

iii) Design a scalable neural network architecture suitable for WSN applications, spanning from small to large scales (including scenarios involving over a hundred nodes).

To validate the performance of the proposed method, we have developed a comprehensive simulation environment based on an actual WSN configuration for SHM. This setup encompasses a total of 112 nodes, installed in the Jindo bridge in South Korea, providing a real-world context for validation and testing.

## 2. System model

This study developed an RL simulation environment that replicates the ways of energy use/consumption, solar energy harvesting, and radio communication of wireless sensors for SHM networks. The Open AI Gym framework was used to create a standardized interface for various training and testing capabilities with the customized simulation environment. In this setup, wireless sensor hardware specifications were taken from the commercial wireless sensor platform Xnode, which was developed based on the Illinois Structural Health Monitoring Project (ISHMP) cc Each wireless sensor node is equipped with a CPU, radio system, vibration sensors, solar charging system, and rechargeable battery. There are two types of nodes in the SHM network: one is a leaf sensor node that actually measures structural responses, listens for events, and transmits the measured data to the gateway node, the other is a gateway node that manages multiple leaf nodes, collects data from them, processes the data, and transmits data to the server. Also, possible uncertainties associated with wireless communication and energy harvesting were considered to mimic realistic WSN operation condition, following the format used in Long and Buyukozturk's study [15].

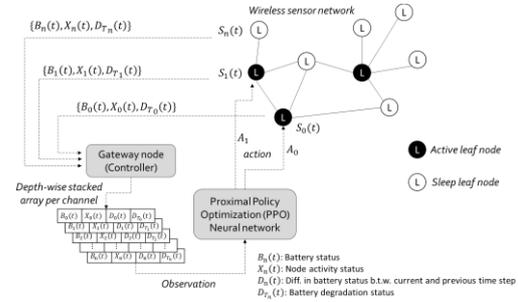

Fig. 1. Proposed system schematic

Fig. 1 shows the schematic of the proposed system where leaf nodes periodically report their battery state to a gateway node. Based on the reported states, the gateway node determines the necessary actions and publishes them to the nodes, leading to changes in the set of active nodes over time. The active nodes maintain local peer-to-peer connections and can trigger a system-wide network awakening from sleep mode when they detect an event of interest. The status of each sensor node is transformed as 2D image format and stacked depth-wisely. It allows to maintain small input size (i.e., single node status array) regardless the number of nodes which need to be handled. This method enables scalable implementation of this DLR method for various node quantity. More detail is described in section 5.

## 2.1 Energy consumption

At regular intervals, the battery reserve of each node in the network is measured and recorded. The time model used in the system is discrete, with each time slice denoted as $t_k$ and having a duration $\Delta t$. The battery reserve of a node $i$ at the beginning of time slice $t_k$ is represented as $B_i(t_k)$. At each period $t_k$, the status of each sensor node is defined as $X_i(t_k)$, where $X_i \in \{0,1,2\}$. The value of $X_i = 0$ signifies that the node is in active mode, $X_i = 1$ indicates the node is in idle mode, and $X_i = 2$ represents that the node is in deep sleep mode. The energy consumption rate of each node depends on its status, with a fixed rate of $P_a$ while active, a reduced rate of $P_i$ while idle, and an even lower rate of $P_s$ while in deep sleep mode.

Table 1. Power consumption of individual sensor node components in active, idle, and deep sleep modes (reproduced from [15])

|  | Active (mW) | Idle (mW) | Deep sleep (mW) |
|---|---|---|---|
| Processor | 203.5 | 1.85 | 0.185 |
| Radio | 166.5 | 166.5 | 0.03 |
| Sensor | 55.5 | 1.85 | 0.185 |
| Total | 425.5 | 170.2 | 0.4 |

In this study, the active energy consumption rate of a sensor node is determined by computing the sum of power consumed by the microcontroller, the radio, and the sensor (i.e., MEMS accelerometer), all in the active state. The energy consumption rate of idle state, denoted as $Pi$, is obtained by combining the power consumed by the idle microcontroller and idle MEMS accelerometer, while the radio remains in an active state, listening for a wake-up signal from a neighboring node. The deep sleep mode, which consumes the least amount of power, is characterized by the radio, processor, and sensor being in their lowest power consumption state, resulting in limited node functionality. The power consumption rates used in this study are specified in Table 1, with a battery capacity of 3,000 mAh and an average discharge voltage of 3.7 V chosen for the simulations. In this study, a flat energy consumption rate was utilized as an example. The off-the-shelf hardware (i.e., Xnode, [11,12]) used has a transmission capability of up to 1km. In the case study, specifically the Jindo bridge WSN configuration, the maximum distance between the gateway node and the leaf node is less than 30% (300m) of the capable range, as depicted in Figure 2. The total energy consumption of each node in each period can be calculated as follows:

$$E_c(t_k) = \begin{cases} P_a \Delta_t \text{ when } X_i(t_k) = 0 \\ P_i \Delta_t \text{ when } X_i(t_k) = 1 \\ P_s \Delta_t \text{ when } X_i(t_k) = 2 \end{cases} \quad (1)$$

## 2.2 Energy harvesting

The battery level at time $t+1$ for sensor $i$, $B_i(t_{k+1})$ can be calculated by considering the energy consumed by the sensor, $E_{c_i}(t_k)$, and the energy harvested by the sensor, $E_{h_i}(t_k)$. The calculation is performed using the following Equation:

$$B_i(t_{k+1}) = B_i(t_k) + E_{h_i}(t_k) - E_{c_i}(t_k) \quad (2)$$

In reality, the harvested energy at each node is subject to spatial variation of the node in a sensor network, even if the energy harvesting capabilities of each node are the same. To account for this effect, we model the spatial variation of energy as a random field over the area of the network. The harvested energy $E_{h_i}(t_k)$ at each sensor $i$ for a time period $t_k$ is given by the baseline mean value for the entire network, $E_\mu(t_k)$, considering a spatially correlated random perturbation $Y_i$ at the location of the sensor $(x, y)$, as shown in Equation 3 and Fig. 2. $Y_i$ is modeled as a zero-mean, stationary, Gaussian random field with an exponential covariance function as shown in Equation 4[16]. The 2D Gaussian random field is simplified to fit within a grid of 4 rows by 30 columns, with a standard deviation ($\sigma$) of 0.01 and a length scale ($l_0$) of 5. This model captures approximately 20-30% variation in the expected value of $E_{h_i}(t_k)$, providing the uncertainty inherent in energy harvesting.

$$E_{h_i}(t_k) = E_\mu(t_k) \cdot (1 + Y_i(x, y)) \quad (3)$$

$$C(x,y) = \sigma^2 \, exp\left(\frac{\|x-y\|}{l_0}\right) \quad (4)$$

The amount of baseline harvested energy $E_\mu(t_k)$ is calculated for our case study location (i.e., Jindo Bridge, Korea) by using National Solar Radiation Database and the System Advisor Model (SAM), 3W solar panel at a 0° tilt for every 30-minute interval located at the Jindo bridge location (i.e., 34.571725ºN, 126.305262ºW). The addition of spatially correlated noise to this baseline time series enhances the realism of the simulation and provides the RL agent with noisy sequences, reducing the risk of overfitting and improving robustness to variations in solar energy availability.

*2.3 Wireless connectivity*

To account for the possible uncertainty associated to the local peer-to-peer radio communication between the sensor nodes, Soft Geometric Graph Model [15,17,18] was considered. The model basically determines the probability of connection between nodes i and j followed by:

$$H_{ij} = \beta exp\left(-\left(\frac{r_{ij}}{r_0}\right)^\eta\right) \quad (5)$$

In this Equation 5, $r_{ij}$ represents the Euclidean distance between nodes $i$ and j, and $r_0$ is the range parameter to describe the signal decay over distance. $\eta$ is an environment-specific parameter, which is assumed to have a constant value of 1 in this study throughout the sensor network. The distance-independent parameter $\beta$, which linearly scales the probability of connection, is also assumed to be 1 for simplicity. The range parameter $r_0$ is set as 1km, considering an existing commercial WSN system(i.e., Xnode, Embedor technologies)'s maximum communication range [19,20]. In this study, we focus solely on the connection between the gateway node and each leaf node. We assume that all sensor data from the leaf nodes are gathered by the gateway node. Changes in the status of each node will only be taken into account when there is successful peer-to-peer communication, as defined by Equation 5.

*2.4 Battery degradation model*

Battery degradation is influenced by various factors such as chemical processes, cycling, temperature, mechanical stress, and manufacturing defects. Among these, charge-discharge cycles and temperature are the primary factors affecting battery degradation. In this study, we focus on charge-discharge cycles as the main variable for controlling battery degradation. This choice is made because typical Wireless Sensor Network (WSN) nodes usually lack specific temperature sensors for batteries. Additionally, we assume that ambient temperature remains consistent across all WSN nodes, given that they operate in the same region under outdoor conditions.

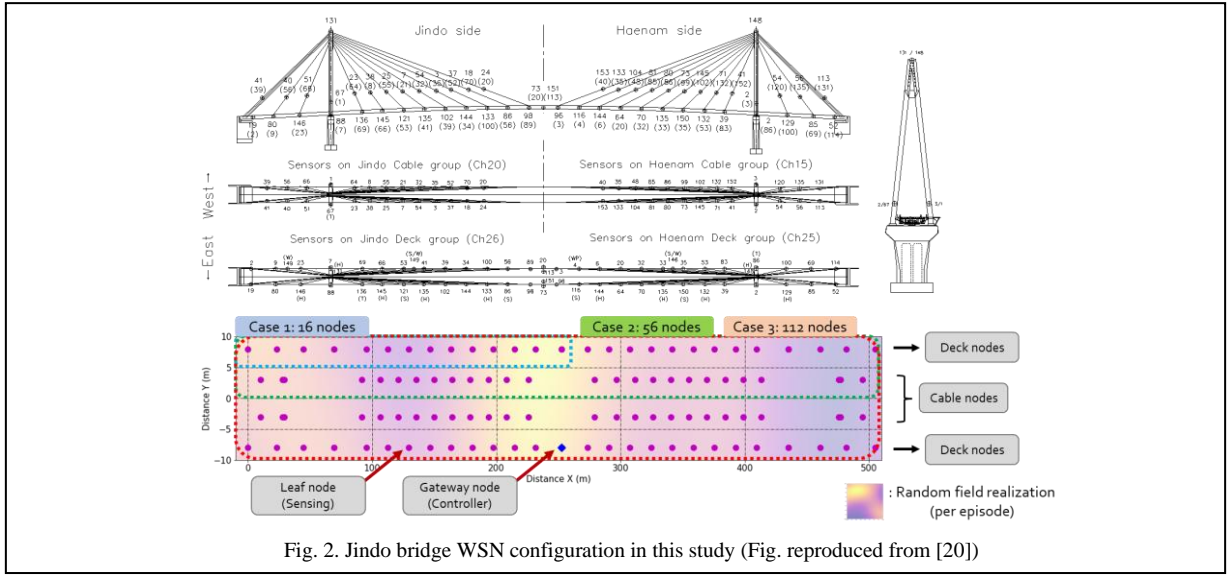

Fig. 2. Jindo bridge WSN configuration in this study (Fig. reproduced from [20])

In this study, a battery degradation model that considers solar photovoltaic variability, battery cycle, and amplitude variation was adopted [21,22].

$$D_T = \sum_i^n \frac{R_i \ Cycles}{A \cdot R_i^B} \cdot 100 (\%) \qquad (6)$$

In this model, total battery degradation $D_T$ for each sensor node is calculated in an accumulated manner by the rainflow counting algorithm [23]. The algorithm considers the depth of charging and discharging, represented by $R_i$, for each cycle (indexed by i), where n is the total number of cycle. The empirical parameters $A$ and $B$ are taken from existing battery specifications for realistic simulation; $A = 3351$ and $B = -1.689$ [22].

*2.5 Test bridge model: Jindo bridges, South Korea.*

The Jindo Bridges are two cable-stayed bridges that connect Haenam, a small town located in the southwest tip of Korea, to Jindo Island. Each bridge is made up of three continuous spans, with a 344-meter central span and two 70-meter side spans. In a previous project, a successful full-scale SHM project using WSN was completed with a total of 112 nodes [24–26]. In this study, three different WSN sizes were used to test the scalability of the method, starting from i) 16 nodes, a part of the network, then increasing the network size to ii) 56 nodes, which is a half of the network, and iii) 112 nodes, which is the entire Jindo bridge WSN. Fig. 2 depicts a drawing of the bridge, WSN configuration, and the groups of sensor nodes used for each case. The gateway node, also known as the controller, is located in the central position of the deck. Although the sensors on the stay cables have different heights, we assumed this setup in 2D plane for simplicity. The random parameter $Y_i$ in Equation (3) is generated every episode.

## 3. Reinforcement learning

Reinforcement learning (RL) is a self-learning framework that relies on experience to guide an agent's training in a given environment. RL problems can be modeled as a Markov decision process (MDP), consisting of states ($s$), actions ($a$), and rewards ($r$). The agent operates within a repeating cycle of discrete time steps, where it observes the environment's state ($s_t$), selects an action ($a_t$), and receives a reward ($r_t$) for the selected action, guiding the environment to transit to a new state ($s_{t+1}$). As long as a problem can be modeled as a MDP, finding an optimal policy that maximizes future rewards based on the current state is possible by using dynamic programming [27]. Also combining deep neural networks and RL has enabled the handling of high-dimensional state spaces, allowing for the application of RL to complex tasks. One successful approach is the deep Q-network (DQN), which combines Q-learning with deep neural networks[28]. However, DQN is limited to low-dimensional action spaces due to the curse of dimensionality, where finer discretization of the action space leads to an exponentially increasing number of actions. To address this limitation, various DRL methods have been developed, such as trust region policy optimization (TRPO) [29], proximal policy optimization (PPO) [30], and deep deterministic policy gradient (DDPG) [31] to handle multi-dimensional continuous/discrete action spaces.

*3.1 Proximal policy optimization (PPO)*

As an on-policy, actor-critic, policy-gradient(PG) approach, PPO is designed to maintain the reliable performance of TRPO algorithms, while employing only first-order approximation; which ensures the monotonic improvements by evaluating the Kullback-Leibler (KL) divergence of policy updates in an efficient way.

The PPO is based on the PG method that has the gradient estimator as follows:

$$\hat{g} = \hat{\mathbb{E}}_t[\nabla_\theta \log \pi_\theta(a_t|s_t) \hat{A}_t] \qquad (7)$$

Where $\pi_\theta$ is a stochastic policy and $\hat{A}_t$ is an estimator for the advantage function at a time step $t$. The estimator $\hat{g}$ is obtained by differentiating the objective function:

$$L^{PG}(\theta) = \hat{\mathbb{E}}_t[log\ \pi_\theta(a_t|s_t)\hat{A}_t] \qquad (8)$$

Although optimizing the loss function $L^{PG}(\theta)$ seems to be simple and intuitive for policy updates, this approach faces several challenges, including sample inefficiency, balancing exploration and exploitation, and high variance in learned policies[30]. In practice, this can lead to excessively large policy updates that can negatively affect future observations and rewards. Advanced policy-based algorithms such as TRPO and PPO have addressed such challenges by employing an actor-critic structure that combines the benefits of both traditional value-based and policy-based methods. The PPO is known to be simpler to implement, requires less computation, and has better sample complexity than TRPO. The PPO uses a clipped surrogate loss function that combines the policy surrogate with a value function error term, as follows:

$$L_t^{CLIP+VF+S}(\theta) = \hat{\mathbb{E}}_t[L_t^{CLIP}(\theta) - c_1 L_t^{VF}(\theta) + c_2 S[\pi_\theta](s_t)] \qquad (9)$$

Where $L_t^{CLIP}$ is the clipped surrogate objective, $c_1$, $c_2$ are coefficients and $S$ denotes an entropy bonus, and $L_t^{VF}$ is a squared-error loss of the value function $(V_\theta(s_t) - V_t^{targ})^2$. Especially, the clipped surrogate objective $L_t^{CLIP}$ takes the form below:

$$L_t^{CLIP}(\theta) = \hat{\mathbb{E}}_t[min(r_t(\theta)\hat{A}_t, clip(r_t(\theta), 1-\varepsilon, 1+\varepsilon)\hat{A}_t)] \qquad (10)$$

In the PPO algorithm, the $r_t(\theta)$ is defined as $r_t(\theta) = \frac{\pi_\theta(a_t|s_t)}{\pi_{\theta_{old}}(a_t|s_t)}$, which is the probability ratio between the current and old policy, where $a_t$ is the action taken by the current policy and $s_t$ is the state. The hyperparameter $\varepsilon$ is used to clip the probability ranges between $1 \pm \varepsilon$ depending on whether the advantage is positive or negative. This clipping process is applied to the advantage approximator $\hat{A}_t$, resulting in the clipped objective. The final value of $L_t^{CLIP}$ is obtained by taking the minimum of this clipped objective and the unclipped objective $r_t(\theta)\hat{A}_t$. This approach effectively prevents excessively large policy updates, compared to the unclipped case.

For the complete PPO algorithm, the trajectory segments have a fixed length of $T$, and $N$ parallel actors gather data at each time step within the $T$-length trajectory. The advantage function is estimated using a truncated form of the generalized advantage estimation algorithm, which is expressed as:

$$\hat{A}_t = \delta_t + (\gamma\lambda)\delta_{t+1} + \cdots + \cdots + (\gamma\lambda)^{T-t+1}\delta_{T-1}$$

$$\text{Where } \delta_t = r_t + \gamma V(s_{t+1}) - V(s_t) \qquad (11)$$

$\gamma$: discount factor

After collecting data at $T$ time steps each, PPO uses this data to construct the loss function defined in Equation (9) and trains it using mini-batch stochastic gradient descent.

## 4. Problem formulation

This section explains the state-space representation and reward function for the proposed system described in section 2. The state representation of each node $i$ during the $t_k$ time period is expressed as:

$$S_i(t_k) = [B_i(t_k), X_i(t_k), D_i(t_k), D_{T_i}(t_k)] \qquad (12)$$

Where, the battery status $B_i(t_k) \in [0, B_{max}]$ is calculated from Equation (2), $X_i(t_k) \in \{0,1,2\}$ indicates the node activity status(i.e., 0: active, 1: idle, and 2: deep sleep), $D_i(t_k) \in [-B_{max}, B_{max}]$ indicates the difference in battery status between the current and previous time step $D_i(t_k) = B_i(t_k) - B_i(t_{k-1})$, and $D_{T_i}(t_k)$ is the battery degradation calculated by equation (6).

The state representation of the WSN system is generated by stacking the individual node's state to make 2D array state representation as shown below:

$$s(t_k) = [S_0(t_k); S_1(t_k); \ldots; S_N(t_k)] \qquad (13)$$

The advantage of this input shaping method can shape the input array for various size of WSN within DRL setup, which is explained in section 5.1. The action space for each node $i$ consists of three discrete action spaces, i.e., $a_i = \{a_{i_0}, a_{i_1}, a_{i_2}\}$, where $a_{i_0}$: active, $a_{i_1}$: idle, and $a_{i_2}$: deep sleep. And the system level action space is given by:

$$A = [a_0, a_1, \ldots a_N] \quad (14)$$

*4.1 Reward function*

The goal of the RL agent in this study is to maximize the node utility meanwhile balancing the battery degradation level over the network. The total reward $r(t_k)$, in a given time period $t_k$, consists of two major parts: i) reward to maximize the overall system utility, and ii) reward to minimize the standard deviation of battery degradation level within the network. The first part of reward $r_1(t_k)$ is defined by using the number of active nodes in given time period $t_k$ denoted as $C_a(t_k)$. The $C_a(t_k)$ is only counted for the active nodes having battery above 825.5 mW (i.e., $P_a$ + 400mW), considering our assumed minimum battery level of 400mW to avoid complete drain, and the nodes having successful communication per Equation (5). This reward for the node utility $r_1(t_k)$ is calculated by dividing the number of active nodes at time $t_k$ by the total number of nodes $N$, and dividing by the maximum number of steps per episode $t_{k_{max}}$. This way allows to consider various network sizes and simulation step numbers, providing a scalable method.

$$r_1(t_k) = \frac{C_a(t_k)}{N \cdot t_{k_{max}}} \quad (15)$$

The second reward component $r_2(t_k)$ takes into account the battery degradation at the system level. It is calculated by finding the standard deviation of $D_T$, which is then normalized by the maximum $D_T$ in $t_k$. This is expressed as:

$$r_2(t_k) = Std\left(\frac{[D_{L_0}(t_k), D_{T_1}(t_k) \ldots D_{T_N}(t_k)]}{D_{T_{max}}(t_k)}\right) \quad (16)$$

$r_2(t_k)$ is also designed to handle various time scale by normalizing each node's degradation level to

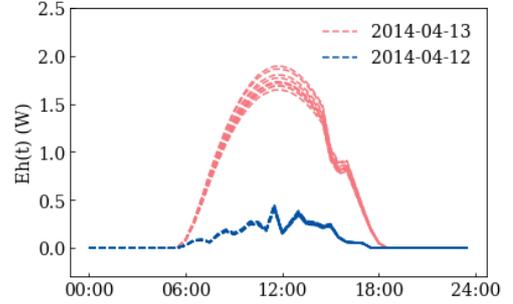

Fig. 3. Harvested solar energy example for two contrasting days

maximum degradation level within a network. Therefore, final reward function is defined as follows:

$$r(t_k) = \alpha_1 r_1(t_k) - \alpha_2 r_2(t_k) \quad (17)$$

For total reward at given time period, reward scaling factors $\alpha_1$ and $\alpha_2$ are used to maximize training efficiency and performance. It is important to keep the rewards within a reasonable range to avoid numerical instability and convergence issues [32]. For example, if the rewards are too large or too small, it may cause numerical overflow or underflow in the computation of gradients, which can make it difficult for the agent to learn. In this study, $\alpha_1$ is set as 6, and $\alpha_2$ is set to 0.05.

## 5. Experimental setup

In this study, a 30-day period (i.e., approximately one month) was used as an episode, and each episode was discretized into equally spaced time steps of $\Delta t_k = 3$ hours, resulting in 240 time steps in a single episode. The four years of solar profile data (2013-2017) described in Section 2.2 were used for training by randomly selecting one month period of data (i.e., 240 steps in a single episode). Then the performance of the trained agent was evaluated using a continuous 12-months period in 2018 (i.e., 240 × 12 = 2880 steps) to validate the long-term performance of the battery degradation balancing control.

The present study involves the generation of a random field during each episode, which serves to provide distinct variations in the solar profile. Fig. 2 shows a random field example in the WSN configuration, while Fig. 3 presents the example of harvested energy in a time series over two distinct days for all nodes.

Table 2. Simulation Scenarios by network size (i.e. node count) and state/reward configuration

| Case | Network size | State | Reward |
|---|---|---|---|
| 1 | 16 | $[B,X,D,D_T]$ | $\alpha_1 r_1 - \alpha_2 r_2$ |
| 2 | 16 | $[B,X,D]$ | $\alpha_1 r_1$ |
| 3 | 56 | $[B,X,D,D_T]$ | $\alpha_1 r_1 - \alpha_2 r_2$ |
| 4 | 56 | $[B,X,D]$ | $\alpha_1 r_1$ |
| 5 | 112 | $[B,X,D,D_T]$ | $\alpha_1 r_1 - \alpha_2 r_2$ |
| 6 | 112 | $[B,X,D]$ | $\alpha_1 r_1$ |

Three WSN configurations having different network sizes (i.e. 16, 56, and 112) have been tested to see the scalability of the method and the effect of $D_T$ within state and reward for battery degradation control performance for each network size were investigated. The details of each case are provided in Table 2. In Cases 1, 3, and 5, each node' state is composed of four components, as shown in Equation (12), and the reward function is given by Equation (17). Cases 2, 4, and 6, however, do not include the battery degradation-related component $D_T$ for a comparison purpose as follows.

$$S_i(t_k) = [B_i(t_k), X_i(t_k), D_i(t_k)] \quad (18)$$

$$r(t_k) = \alpha_1 r_1(t_k) \quad (19)$$

*5.1 Neural network architecture*

The neural network architecture in this study consists of a convolutional neural network (CNN)-based feature extractor, action network, and value network, as shown in Table 3. The shape of each node's state was represented by state array starting from input shape (i.e., 4×1 array), and gradually increased to 20×17 by adding zero padding of size 2 throughout the eight CNN layers. These padded arrays were then stacked by the number of nodes vertically, resulting in an array shape of [node counts, height, width, 1]. The size of the feature map in the neural network was selected in a power of two, making the feature size just bigger than each network size. In other words, for Case 1 and 2, which had 16 nodes, a feature size of $2^5 = 32$ was used. For Case 3 and 4, which had 56 nodes, a feature size of $2^6 = 64$ was used. For Case 5 and 6, which had 112 nodes, a feature size of $2^7 = 128$ was used. Each node's state was directly input into the feature map, allowing the input image size to be maintained regardless of the number of sensor nodes in the WSN. The convolutional layers had a kernel size of 3×3 with stride (1,1), and the Leaky-Relu activation function was used, while the linear layers had a Tanh activation function. At the end of feature extractor network, global average pooling layer and linear were used to flatten network dimension. The action network and value network both consist of two linear layers with the same feature size. The action network directly produces node status output to alter node status, while the value network calculates the value of the state.

Table 3. Neural network architecture

| Input | State | |
|---|---|---|
| Feature extractor | Conv. 2D [feature: 32] ×8 | |
| | Global average pooling layer | |
| | Linear [feature: 32] | |
| | **Action net.** | **Value net.** |
| NN | Linear [feature: 32] ×2 | Linear [feature: 32] ×2 |
| Output | [node count ×3] | 1 |

*5.2 Training and test setup*

The training was conducted using a batch size of 960, which included four episodes for each network update. The learning rate was set to 0.00003 and a discount factor was not used as the problem had a fixed time frame. The clipping range was set to 0.2, the clipping parameter for the value function was set to 0.5, and the generalized advantage estimator (GAE) parameter was set to 0.95. The training was performed on the Ocelote GPU node equipped with Intel Haswell/Broadwell 28 core processors, 168 GB RAM, and NVIDIA P100 16 GB GPU at the University of Arizona.

## 6. Results

In this section, the results of a proposed WSN control method using the PPO agent, trained with 20,000 episodes, are presented. Fig. 4 shows the learning curve example for Case 1, which demonstrates that the PPO agent can learn to maximize the reward, i.e., increasing WSN utility while balancing battery degradation level in a network/system level. After training with a 30-days period, the PPO agent was tested with a 12-months continuous period using the 2018 test dataset. Table 4 presents the average and standard deviation of the active node count, allowing for quantitative comparison between cases with and without the battery degradation model $D_T$ in state and reward for three different WSN setups. Fig. 5 visualizes the performance of each case. The top section of each Fig. displays the battery degradation level, along with the counts of active, idle, and sleep mode instances for every node throughout a one-year testing period. A distinct contrast is noticeable between cases that account for battery degradation (i.e., cases 1, 3, and 5) and those that do not (i.e., cases 2, 4, and 6). The incorporation of a reward system aimed at minimizing the standard deviation of battery degradation has evidently played a role in enhancing the maximization of system utility. Moreover, this approach has contributed to a more even distribution of battery degradation across the entire network. Overall, Cases 1, 3, and 5, which considered the battery degradation model, showed around 90% of system utility during the year, allowing similar battery degradation across the WSN as shown in Table 4. In contrast, cases 2, 4, and 6 that did not include the battery degradation model showed significantly less system utility and imbalanced battery degradation performance compared to Cases 1, 3, and 5. Such behaviors can be observed from Fig. 5 as well; while uniform behaviors are obtained for Case 1, 3, and 5 in terms of the total battery degradation $D_T$ and the number of active/idle/sleep modes of each sensors over the testing period of 1 year, but quite dispersed behaviors were observed for Case 2, 4, and 6. This result indicates that employing the battery degradation model $D_T$ in the method (i.e., Case 1, 3, and 5) can effectively maximize system utility, while ensuring

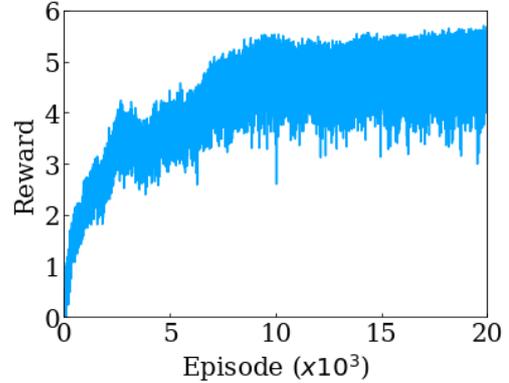

Fig. 4. Learning curve example (Case 1)

balanced use of sensor batteries in a system level to prevent imbalanced battery degradation.

The proposed method for shaping input data involves stacking each node's state array and directly feeding it to a feature map. This method prevents a drastic increase of input size and enables the model to maintain its performance by increasing the feature map size in proportion to the number of nodes. However, when the network size increased, cases without the battery degradation model showed worse performance for all cases. This highlights the importance of the battery degradation model in ensuring optimal system performance.

The lower part of each Fig. within Fig. 5 shows the average node count in the time domain with a solar energy harvesting profile. Cases 1, 3, and 5 showed very similar trends, attempting to maximize system utility over time. In contrast, Cases 2, 4, and 6 attempted to maintain the overall WSN utility at a constant number. Cases 1, 3, and 5 showed somewhat fluctuation in the active node count over 1 year period, but it can be explained by the fluctuation of the harvested solar energy as shown by the grey bar in the Fig. 5 (i.e., $E_h(t)$ ($W$) in right axis). In contrast, Cases 2, 4, and 6 display smaller fluctuations in the active node count, suggesting that the agent switches nodes on/off to maintain a specific number of active nodes consistently. This behavior aligns with findings from the study by Long and Buyukozturk, where certain nodes are activated more frequently, some act as reserve nodes when active nodes are unavailable, and others remain in sleep mode [15]. This approach may be usable for small WSNs, where more than half of the nodes are predominantly active (e.g., Case 1 & 2). However, in larger networks, such as Cases 3-6, a considerable number of active nodes are lost over time, which is impractical. This result clearly demonstrates that the proposed method has the advantage of maximizing system utility regardless of the WSN size, while also effectively managing battery degradation at the system level.

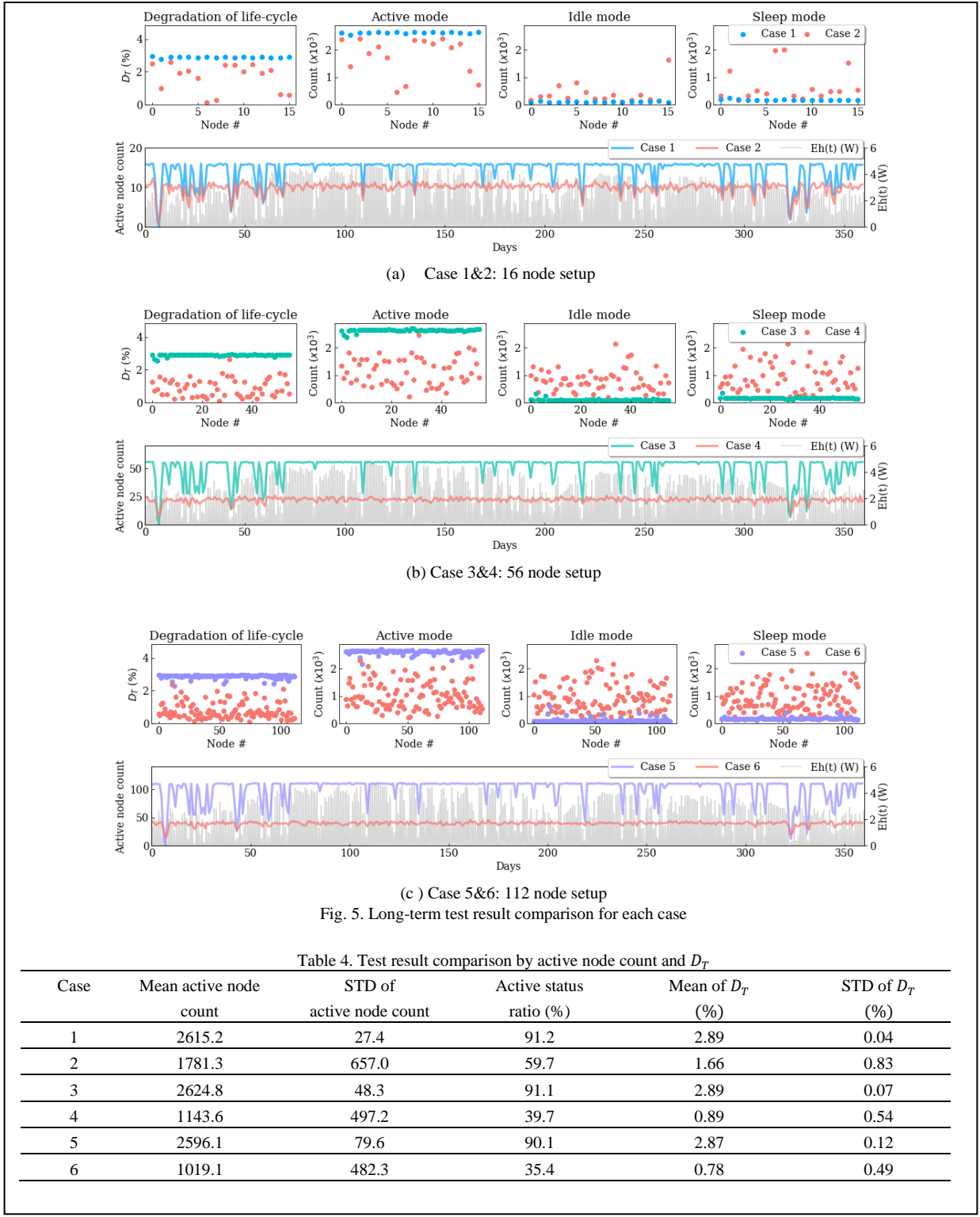

(a) Case 1&2: 16 node setup

(b) Case 3&4: 56 node setup

(c) Case 5&6: 112 node setup

Fig. 5. Long-term test result comparison for each case

Table 4. Test result comparison by active node count and $D_T$

| Case | Mean active node count | STD of active node count | Active status ratio (%) | Mean of $D_T$ (%) | STD of $D_T$ (%) |
|---|---|---|---|---|---|
| 1 | 2615.2 | 27.4 | 91.2 | 2.89 | 0.04 |
| 2 | 1781.3 | 657.0 | 59.7 | 1.66 | 0.83 |
| 3 | 2624.8 | 48.3 | 91.1 | 2.89 | 0.07 |
| 4 | 1143.6 | 497.2 | 39.7 | 0.89 | 0.54 |
| 5 | 2596.1 | 79.6 | 90.1 | 2.87 | 0.12 |
| 6 | 1019.1 | 482.3 | 35.4 | 0.78 | 0.49 |

In summary, Cases 2, 4, and 6 exhibit lower overall system utility performance over time, while the PPO agent trained in Cases 1, 3, and 5 effectively maximizes system utility while managing battery degradation across three different WSN setups.

## 7. Conclusion

In this study, we proposed a DRL based approach for autonomous control of WSN that aims to maximize system utility and balance the battery degradation level of sensor nodes. To ensure the realistic simulation of practical conditions, we considered a real-world WSN setup (Jindo bridge WSN) for SHM, possible uncertainties associated to wireless communication, spatial randomness of sensor nodes within geospatial condition, and real-life solar energy profile. We adopted a battery degradation model to manage and centralized control the sensor nodes' battery degradation at the WSN system level. We also proposed a new neural network architecture design for scalable implementation to various network sizes. Additionally, we proposed a state array shaping method that depth-wise stacks each node's state array to maintain the input array's shape for feeding into the neural network, but feed each node's state into feature map dimension directly.

The proposed method's effectiveness was confirmed through a year-long operation of the WSN. The trained agent successfully optimized system utility while also managing battery degradation within the network over the entire year. Comparative studies highlighted that integrating the battery degradation model significantly enhanced the agent's performance by maximizing system utility and maintaining battery health in the WSN.

Future research should focus on improving the selection of active nodes by incorporating data quality factors, such as Modal Assurance Criterion (MAC) analysis. Additionally, it should develop more robust battery degradation models to accurately predict realistic battery degradation trends, taking into account various influencing factors. This approach could potentially reduce overall energy consumption by optimizing node selection at each step, thereby extending the overall battery life of the WSN system.

## Acknowledgments

This study was supported by the United States National Science Foundation (NSF) under award number CMMI 2027425.